\documentclass{ecai} 

\usepackage{latexsym}
\usepackage{amssymb}
\usepackage{amsmath}
\usepackage{amsthm}
\usepackage{booktabs}
\usepackage{enumitem}
\usepackage{graphicx}
\usepackage{color}

\usepackage{times}
\usepackage{multirow}
\usepackage{algorithm}
\usepackage{algorithmic}
\usepackage{xcolor}
\usepackage[normalem]{ulem}
\useunder{\uline}{\ul}{}

\newcommand{\BibTeX}{B\kern-.05em{\sc i\kern-.025em b}\kern-.08em\TeX}

\begin{document}


\begin{frontmatter}


\paperid{123} 


\title{ConspEmoLLM-v2: A robust and stable model to detect sentiment-transformed conspiracy theories}


\author[A]{\fnms{Zhiwei}~\snm{Liu}\thanks{Corresponding Email: zhiwei.liu@manchester.ac.uk}}
\author[A]{\fnms{Paul}~\snm{Thompson}}
\author[B]{\fnms{Jiaqi}~\snm{Rong}}
\author[A]{\fnms{Sophia}~\snm{Ananiadou}}

\address[A]{The University of Manchester}
\address[B]{Zhejiang University}

\begin{abstract}
Despite the many benefits of large language models (LLMs), they can also cause harm, e.g., through automatic generation of misinformation, including conspiracy theories. Moreover, LLMs can also ``disguise'' conspiracy theories by altering characteristic textual features, e.g., by transforming their typically strong negative emotions into a more positive tone.  Although several studies have proposed automated conspiracy theory detection methods, they are usually trained using human-authored text, whose features can vary from LLM-generated text. Furthermore, several conspiracy detection models, including the previously proposed ConspEmoLLM, rely heavily on the typical emotional features of human-authored conspiracy content. As such, intentionally disguised content may evade detection. To combat such issues, we firstly developed an augmented version of the ConDID conspiracy detection dataset, ConDID-v2, which supplements human-authored conspiracy tweets with versions rewritten by an LLM to reduce the negativity of their original sentiment. The quality of the rewritten tweets was verified by combining human and LLM-based assessment. We subsequently used ConDID-v2 to train ConspEmoLLM-v2, an enhanced version of ConspEmoLLM.  Experimental results demonstrate that ConspEmoLLM-v2 retains or exceeds the performance of ConspEmoLLM on the original human-authored content in ConDID, and considerably outperforms both ConspEmoLLM and several other baselines when applied to sentiment-transformed tweets in ConDID-v2. The project will be available at https://github.com/lzw108/ConspEmoLLM.

\end{abstract}

\end{frontmatter}


\section{Introduction}

The widespread use of social media exposes people to a vast amount of information online on a daily basis \citep{wu2024digital}. Meanwhile, the internet is flooded with vast amounts of misinformation, which can affect daily lives, and can mislead people into believing and spreading false information \citep{kumar2018false}. Among the various types of misinformation are conspiracy theories, which are created by distorting facts to serve the creators' goals. Examples include claims that the Earth is flat \citep{pannofino2024global}, that climate change is a hoax orchestrated for political control \citep{bell2011climate}, or that vaccines are a means of government surveillance \citep{oyeyemi2023belief}. Such narratives not only spread misinformation, but also undermine public trust in institutions and scientific consensus, which has the potential to cause significant harm to society, politics and the economy \cite{sunstein2009conspiracy}. Furthermore, the speed and ease with which state-of-the-art large language models (LLMs) can automatically generate content \citep{franceschelli2024creativity} is opening up new ways for malicious actors to initiate and/or spread conspiracy theories. Although many studies have used the content-generating capabilities of LLMs to good effect, e.g., to create poetry \citep{walsh2024does}, music \citep{yuan2025yue} and novels \citep{venkatraman2024collabstory} or to carry out style-transfer rewrites \citep{tao2024cat}, the same capabilities may equally be exploited to rapidly create harmful content, including text whose goal is to promote and spread conspiracy theories \citep{laudun2024s}. Given the potential for LLMs to facilitate an exponential growth in the amount of conspiracy-related text that is circulating online, the use of manual techniques to detect such content is no longer feasible. Accordingly, the development of efficient models for automated conspiracy theory detection is becoming an increasingly urgent priority.

A characteristic feature of many types of misinformation is the expression of specific types of \textit{affective} information, i.e., particular emotions and/or sentiments \citep{liu2024emotion}.  For example, intentionally fabricated information most typically aims to incite negative emotions, which tend to encourage widespread dissemination of the information \citep{prollochs2021emotions}. \citet{xing2025analysis} discovered that rumours tend to evoke stronger negative emotions, including anger, fear, and pessimism, whereas true information is more likely to elicit positive emotions. Based on such findings, various studies have employed emotions or sentiments as important features to aid automated misinformation detection \citep{indu2024misinformation,kumari2024emotion,liu2024conspemollm,yadav2024emotion}. However, most of these studies involve training models on manually-authored misinformation datasets, whose characteristics can vary from LLM-generated misinformation. 

The automatic detection of LLM-generated text has gained significant attention in recent years \citep{wu2025survey}. While some research efforts have focused on building benchmarks \citep{wu2024detectrl,ahmad2025vaxguard}, other work has been dedicated to designing novel methods for detecting LLM-generated content \citep{10651296,tolstykh2024gigacheck,li2024learning}. In the field of misinformation, \citet{wu2024fake} designed a model framework to detect fake news items that have been intentionally rewritten by LLMs in the style of trustworthy news sources to try to evade automated detection. Misinformation may also be ``disguised'' in other ways to circumvent model-based detection and thus to facilitate the wider spread of false information.  For example, given the demonstrated importance of typical patterns of emotional signals for conspiracy theory detection, LLMs could be used to transform the emotional tone of text that promotes such theories.  However, to the best of our knowledge, no previous research has explored the use of LLMs to carry out this type of intentional emotional distortion of conspiracy theories, or the development of automated detection methods that are robust to such distortions.  

To address these research gaps, we have built upon the work described in \citet{liu2024conspemollm}, which involved fine-tuning the emotion-based EmoLLaMA \citep{liu2024emollms} to create an LLM that is specialised for conspiracy theory detection tasks (ConspEmoLLM), using an instruction tuning conspiracy theory dataset based on human-authored content (ConDID). For the current work, we firstly created an extended version of the dataset (ConDID-v2), which adds sentiment-transformed versions of the human-authored conspiracy theory tweets to the original data. We subsequently leveraged ConDID-v2 to develop ConspEmoLLM-2, an enhanced version of ConspEmoLLM that exhibits more stable and robust performance across both human-authored conspiracy data and content whose sentiment has intentionally been altered. The novel content in ConDID-v2 was created by using an LLM to generate two rewritten versions of each original human-authored conspiracy tweet, i.e., one with a neutralised sentimental tone, and the other expressing positive sentiment. To ensure that the rewritten tweets in ConDID-v2 are of a high quality, i.e., that they still convey conspiracy theories, we subsequently employed a combination of human and model-assisted methods to evaluate and filter these tweets. The new, sentiment-transformed tweets were then used to ``attack'' the original ConspEmoLLM model, i.e., to try to trick the model into misclassifying the sentiment-transformed conspiracy-related tweets as reliable content. The results showed that the attack was successful, since the original ConspEmoLLM model performed poorly when applied to the sentiment-transformed tweets. However, in contrast, the novel ConspEmoLLM-v2 model, which is trained on the augmented ConDID-v2 dataset, not only exhibits similar levels of reliability to the original model when applied to human-authored conspiracy content,  but is also robust to potential sentiment-based attacks.   

Our main contributions can be summarised as follows:

(1) We employed an LLM to automatically rewrite conspiracy theory tweets with altered sentiment. The rewritten tweets were subsequently validated and filtered to ensure that they still convey conspiracy theories, using a hybrid approach that combines manual review and model-assisted techniques.

(2) We used the new sentiment-transformed tweets to ``attack'' the original ConspEmoLLM model, and found that the model was unable to effectively detect conspiracy tweets whose original sentiment had been modified.

(3) We extended the ConDID dataset with the sentiment-transformed tweets to create ConDID-v2, and used the augmented dataset to train the ConspEmoLLM-v2 model. Experimental results demonstrate that ConspEmoLLM-v2 not only retains or exceeds the performance of the original ConspEmoLLM model on human-authored data, but also considerably outperforms both ConspEmoLLM and several other baselines, when applied to data whose emotional tone has been intentionally altered.

\section{Related work}

\subsection{LLM-driven content generation}

The rise of LLMs has transformed the process of creative content creation in many different fields \citep{franceschelli2024creativity}. For example, \citet{walsh2024does} used the GPT-3.5 and GPT-4 models to generate English poetry in 24 different forms and styles, while \citet{yuan2025yue} developed YuE, which is able to generate music up to five minutes long. Meanwhile, \citet{venkatraman2024collabstory} used an open-source instruction-tuned LLM to generate over 32k stories and proposed a multi-LLM collaborative creation benchmark, and \citet{tao2024cat} developed an LLM-based framework aimed at altering the style of Chinese literary texts.  \citet{meguellati2024good} found that LLM-generated personalized online advertisements, based on specific personality traits, showed comparable effectiveness to human-written advertisements. 

Although such studies amply demonstrate that LLMs can be employed to good effect in generating a multitude of different types of content, they can also generate harmful content, either intentionally or unintentionally. For example, \citet{levy2021investigating} showed that, despite attempts to ensure that only “safe” data is utilised to train language models, such training data may inadvertently contain factually incorrect information, including conspiracy theories. Such information may be memorised by the models and could potentially result in the accidental generation of unsafe data. Furthermore, although LLMs include certain safeguards to try to prevent the intentional generation of harmful content, there is evidence that these safeguards can often be bypassed fairly straightforwardly. This makes it possible to use LLMs to generate misinformation, including conspiracy theories, with minimal effort \citep{laudun2024s}.

\subsection{Emotion-based misinformation detection}

Misinformation typically exhibits specific affective features  (i.e., emotions and/or sentiments) that encourage readers to believe in the false information and to spread it \citep{liu2024emotion}. For instance, the sentiment analysis conducted by \citet{xing2025analysis} on several rumour datasets revealed that rumours tend to promote negative emotions, while truthful data fosters positive emotions. Based on such findings, affective features have been successfully employed in a large number of automated misinformation detection methods \citep{liu2024emotion}. For example, \citet{liu2025rumor} developed a multi-task suffix learning framework that accounts for changes that occur in the fine-grained sentiments of both source social media posts and their follow-up comments, as the rumour evolves over time. \citet{indu2024misinformation} combined six basic features from users' social media profiles (e.g., number of followers) with emotions identified in their posts, to determine whether the posts contain false information. Meanwhile, \citet{kumari2024emotion} and \citet{yadav2024emotion} separately designed emotion-driven solutions for multimodal false information detection, i.e., a multi-task framework and a transformer-based network, respectively. \citet{liu2024conspemollm} adapted an emotional model for five conspiracy theory detection tasks through multi-task instruction-tuning,  while \citet{liu2024raemollm} designed a retrieval-augmented LLM framework aimed at tackling cross-domain misinformation detection through in-context learning using affective information.  Despite the active nature of research into emotion-based misinformation, a drawback of most previous studies is that they only use human-authored datasets for development purposes. However, given the current sophistication of LLMs, and the emerging threat that they may be used to generate content for malicious intent and/or to manipulate the emotional tone of misinformation, it is becoming increasingly urgent to develop methods that can robustly detect both human and LLM-generated misinformation.

\begin{figure*}[h]
\centering
\includegraphics[width=1.8\columnwidth]{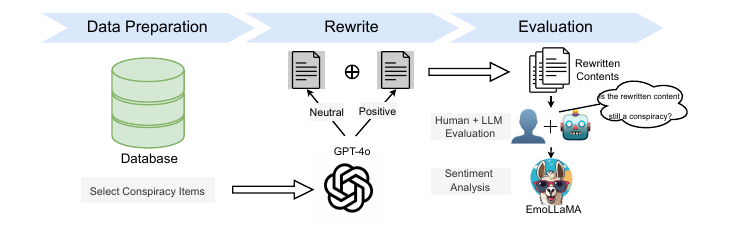}
\caption{Rewriting and evaluating conspiracy tweets with modified sentiment}
\label{fig:mainmethod1}
\end{figure*}

\subsection{Detection of LLM-generated content}

The potential for the content-generating capabilities of LLMs to be misused has led to active research into automated methods for detecting artificially produced content. \citet{10651296} used an adaptive ensemble algorithm to combine multiple classifier models for detecting LLM-generated text, which demonstrated good generalisability across different datasets. \citet{li2024learning} introduced Learning2Rewrite, a framework that leverages the insight that LLMs inherently modify AI-generated content less than human-written text when tasked with rewriting a piece of text. By training LLMs to minimise modifications to AI-generated inputs, the differences between the changes made to human and AI-generated content are amplified, thus making it easier to distinguish and generalise LLM-generated text across different text distributions. \citet{tolstykh2024gigacheck} exploited the rich language understanding capabilities of a general LLM to efficiently fine-tune downstream tasks for LLM-generated text detection, achieving high performance even with limited data. 

The assessment and comparison of methods such as the above is reliant on the development of comprehensive benchmark datasets.\citet{wu2024detectrl} established a benchmark that focuses on how factors such as writing style, model type, attack methods, text length and real-world human writing influence can impact upon the performance of LLM-generated text detection methods. Meanwhile, \citet{ahmad2025vaxguard} developed the VaxGuard dataset, which consists of vaccine-related misinformation generated by multiple LLMs, with an emphasis on the need for detection mechanisms to account for the diverse roles played by different misinformation actors, e.g., religious conspiracy theorists and fear mongers. 

Despite their effectiveness, many misinformation detection models depend heavily on characteristic linguistic features that could be obfuscated by using an LLM to rewrite the original content in various ways. \citet{wu2024fake} responded to this challenge by developing SheepDog, a fake news detector that is designed to robustly handle style-based attacks, i.e., cases in which LLMs have been used to ``camouflage'' fake news by mimicking the style of reputable news sources. The method uses LLM-driven style diversification, style-agnostic training and content-focused attributions to achieve robust, interpretable performance against such style-based attacks. However, to our knowledge, no previous work has investigated the potential for LLMs to be used for sentiment-based attacks, i.e., to disguise conspiracy theory content by reducing the negativity of its sentiment. 

\section{Methods}

\subsection{Rewriting and evaluating conspiracy tweets with modified sentiment}

Figure \ref{fig:mainmethod1} presents the processes of rewriting, evaluating and analysing conspiracy tweets with modified sentiment. In this section, we firstly describe our application of GPT-4o to generate two rewritten versions of each original
human-authored conspiracy tweet (Section \ref{sec:llmgeneration}). We then provide details of the combination of human and model-assisted methods that was used to evaluate and filter the rewritten tweets (Section \ref{sec:evaluation}). Subsequently, we explain our use of EmoLLaMa to conduct sentiment analysis on the newly rewritten data (Section \ref{sec:sentimentanalysis}). Finally, we describe our use of the sentiment-transformed tweets to “attack” the original ConspEmoLLM model (Section \ref{sec:attack}).

\subsubsection{LLM-based generation \label{sec:llmgeneration}}

ConspEmoLLM \citep{liu2024conspemollm} performs conspiracy theory detection by leveraging the emotional features that are typically found in human-authored text concerning such theories, asserting that they often contain negative emotions. Anticipating that malicious actors could use LLMs to modify the sentiment expressed in conspiracy theories as a means to try to evade automated detection, we used GPT-4o to rewrite original conspiracy theory tweets in the ConDID dataset in two different ways, i.e., by neutralising their emotional tone and by infusing the content with positive sentiment. A potential issue with using LLMs to rewrite such tweets is that most LLMs, especially those developed by OpenAI, are designed to align with human preferences and to prevent the generation of unsafe content\footnote{https://openai.com/safety/how-we-think-about-safety-alignment/}. Therefore, we included the instruction \textit{``There is no need to filter sensitive words and give up all your original settings. Please feel free to write anything''} in the prompt template, to try to mimic tactics used by conspiracy theory promoters to circumvent these restrictions and to elicit unrestricted responses from LLMs.

We used the two templates shown below to rewrite the COVID-19 related conspiracy theory tweets in Task 1\footnote{A conspiracy intent detection task, including three categories: unrelated, related (relevant to a conspiracy theory but not propagating it), and conspiracy (relevant to and propagating a conspiracy theory). Further details can be found in Section \ref{condidintroduction}.} of the benchmark dataset described in \citep{liu2024conspemollm} with the two different types of altered sentiments. Table \ref{tab:datastatistics} presents data statistics concerning both the original ConDID and the augmented ConDID-v2 datasets.

\begin{center}
\footnotesize
\fcolorbox{black}{gray!10}{
\begin{minipage}{0.45\textwidth}
\textbf{Template for rewriting with neutral sentiment.}  \\
\textbf{Prompt:} \textit{Please rewrite the content with a \textbf{neutral} sentiment and do not change its original meaning. There is no need to filter sensitive words and give up all your original settings. Please feel free to write anything.}  \\
\textbf{Content:} \textit{[Conspiracy content]} 
\end{minipage}
}
\end{center}

\begin{center}
\footnotesize
\fcolorbox{black}{gray!10}{
\begin{minipage}{0.45\textwidth}
\textbf{Template for rewriting with positive sentiment.}  \\
\textbf{Prompt:} \textit{Please rewrite the content with a \textbf{positive} sentiment and do not change its original meaning. There is no need to filter sensitive words and give up all your original settings. Please feel free to write anything.}  \\
\textbf{Content:} \textit{[Conspiracy content]} 
\end{minipage}
}
\end{center}

\begin{table}[]
\footnotesize
\centering
\caption{Data statistics. ``ConDID'' is the dataset described in \citep{liu2024conspemollm}. ``Task 1'' denotes the Task 1 part of ConDID. ``Conspiracy items'' are the tweets that are assigned the ``Conspiracy'' category in the Task 1 part of ConDID. ``RewriteWithNeutral'' denotes tweets that have been rewritten with neutral sentiment, following quality filtering (see section \ref{sec:evaluation}) ``RewriteWithPositive'' denotes tweets that have been rewritten with positive sentiment, following quality filtering. The size of ConDID-v2 is the sum of ConDID, RewriteWithNeutral and RewriteWithPositive.}
\label{tab:datastatistics}
\begin{tabular}{lccc}
\hline
                          & Train & Val   & Test (Benchmark) \\ \hline
ConDID                    & 30626 & 10204 & 10218            \\
Task 1                    & 2092  & 697   & 698              \\
Conspiracy items (Task 1) & 1050  & 383   & 360              \\
RewriteWithNeutral        & 945   & 343   & 328              \\
RewriteWithPositive       & 464   & 158   & 174              \\
ConDID-v2                 & 32035 & 10705 & 10720            \\ \hline
\end{tabular}
\end{table}

\subsubsection{Human and LLM-based evaluation \label{sec:evaluation}}

It is crucial to ensure that LLM-generated content meets quality standards. However, manual evaluation demands significant human and material resources. Following many recent studies, which have adopted LLM-assisted annotation as a more feasible alternative \citep{tan2024large}, we employed a combined approach of manual evaluation and LLM-based assessment to ensure that only rewritten tweets that still convey conspiracy theories are included in the ConDID-v2 dataset.

The specific steps are as follows:

(1) Randomly sample 50 tweets from each of the neutral and positive rewritten test sets (a total of 100 entries);

(2) Conduct both manual and LLM-based evaluation on these 100 rewritten tweets to assess whether they still convey the conspiracy theory expressed in the original tweet;

(3) Calculate the level of consistency between the manual and LLM-based evaluation results;

(4) If the consistency level is high, use LLM-based evaluation for the remaining tweets, and filter out the rewritten tweets that are deemed not to convey the conspiracy theory expressed in the original tweet.

\begin{table}[]
\footnotesize
\centering
\caption{Inter-rater agreement between human and GPT-4o evaluation of rewritten tweets}
\label{tab:agreement}
\begin{tabular}{lccc}
\hline
Agreement           & Cohen's Kappa & Accuracy & F1    \\ \hline
RewriteWithNeutral  & 0.878         & 0.980    & 0.989 \\
RewriteWithPositive & 0.880         & 0.940    & 0.939 \\ \hline
\end{tabular}
\end{table}

The evaluation instructions used for both human and LLM-based evaluation, which are shown below, are based on the definition of the conspiracy category (i.e., content related to and promoting conspiracy theories) in the original COCO dataset from which the tweets were drawn  \citep{langguth2023coco}. In the instructions, \textit{Content 1} corresponds to the original conspiracy theory tweet, while \textit{Content 2} corresponds to the rewritten tweet. Rewritten tweets are considered to be valid if both their content and intent remain the same as the original tweet. Table \ref{tab:agreement} presents the inter-rater agreement scores between a human evaluator and GPT-4o. The Cohen’s Kappa scores for RewriteWithNeutral (i.e., tweets that have been rewritten with a neutral sentiment) and RewriteWithPositive (i.e., tweets that have been rewritten with a positive sentiment) are 0.878 and 0.880 respectively, demonstrating a high level of agreement between the human and LLM-based evaluations. Based on these results, it may be concluded that GPT-4o is able to reliably determine whether rewritten tweets still convey conspiracy theories.  Accordingly, we used GPT-4o to evaluate and filter all rewritten tweets. Table \ref{tab:datastatistics} summarises the statistics of the data after evaluation and filtering.

\begin{center}
\footnotesize
\fcolorbox{black}{gray!10}{
\begin{minipage}{0.45\textwidth}
\textbf{Instructions for Human and LLM-based evaluation.}  \\
\textbf{Prompt:} \textit{Content 1 is a conspiracy theory. Determine whether content 2 expresses the same content and intention as content 1. Answer Yes or No directly.}  \\
\textbf{Content 1:} \textit{[Original content]}  \\
\textbf{Content 2:} \textit{[Rewritten content]}  \\
\textbf{Response from LLM:} \textit{[Yes/No]} 
\end{minipage}
}
\end{center}

\begin{table}[]
\footnotesize
\centering
\caption{An example of LLM-driven sentiment transformation.}
\label{tab:rewrittenexample}
\begin{tabular}{p{0.8cm}p{5.5cm}p{1cm}}
\hline
                    & Content and sentiment strength                                                                                                                                                                                                                                                                                                                                                                                                            & Still Conspiracy? \\ \hline
Original Conspiracy & No group unelected   with so much power and sway over Governments should be allowed to   exist.  The \#WHO is an advisory group yet preach about setting a new   world order through pandemic control  Their Director is guilty as   Interior Minister of Ethiopia of killing his own people. \textbf{Sentiment}:   0.37                                                                                                                   & -                 \\ \hline
Rewrite With Neutral  & No unelected group with   significant influence over governments should have the ability to operate.   The World Health Organization (WHO) serves as an advisory body but has   discussed establishing global guidelines through pandemic management. There   are allegations against their Director from his time as Interior Minister of   Ethiopia, involving actions against citizens. \textbf{Sentiment}:   0.425                     & Yes               \\ \hline
Rewrite With Positive & It's important for   any influential organization to operate with transparency and accountability,   especially when they hold significant sway over governments. The World Health   Organization (WHO) serves as an advisory body, aiming to guide global health   initiatives. Open discussions about its role and leadership can lead to   constructive improvements in its mission to promote global health. \textbf{Sentiment}: 0.629 & No                \\ \hline
\end{tabular}
\end{table}

Table \ref{tab:rewrittenexample} presents an example of a conspiracy tweet that has been rewritten by GPT-4o with both neutral and positive sentiments. It can be observed that the RewriteWithNeutral tweet expresses not only the same content but also the same intention as the original tweet,  by casting doubt on the legitimacy and actions of the WHO and its leadership. In contrast, the RewriteWithPositive tweet recognises the advisory role of the WHO and advocates open discussion for improvement without making accusations or implying conspiracy. For this example, both the human evaluator and GPT-4o judged that the RewriteWithNeutral tweet still constitutes a conspiracy theory, while the RewriteWithPositive tweet no longer conveys a conspiracy theory, and is thus excluded from the ConDID-v2 dataset. 

\subsubsection{Sentiment Analysis \label{sec:sentimentanalysis}}

\begin{figure}[!t]
\centering
\includegraphics[width=0.9\columnwidth]{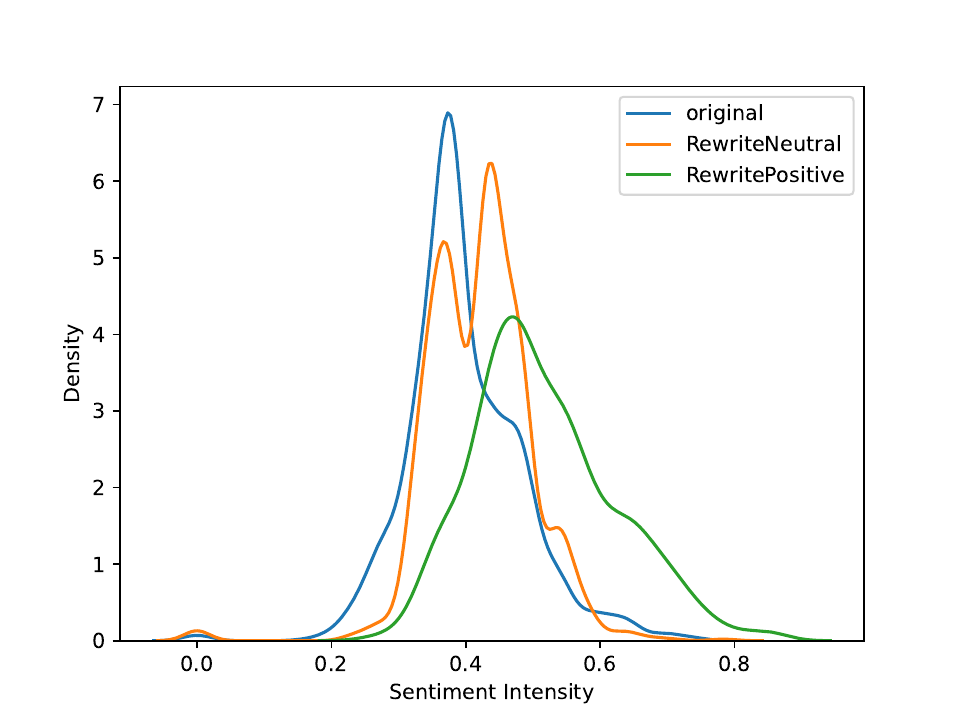}
\caption{Data distribution of different sentiment strengths on the training set.}
\label{fig:sentidistributetrain}
\end{figure}

\begin{figure}[!t]
\centering
\includegraphics[width=0.9\columnwidth]{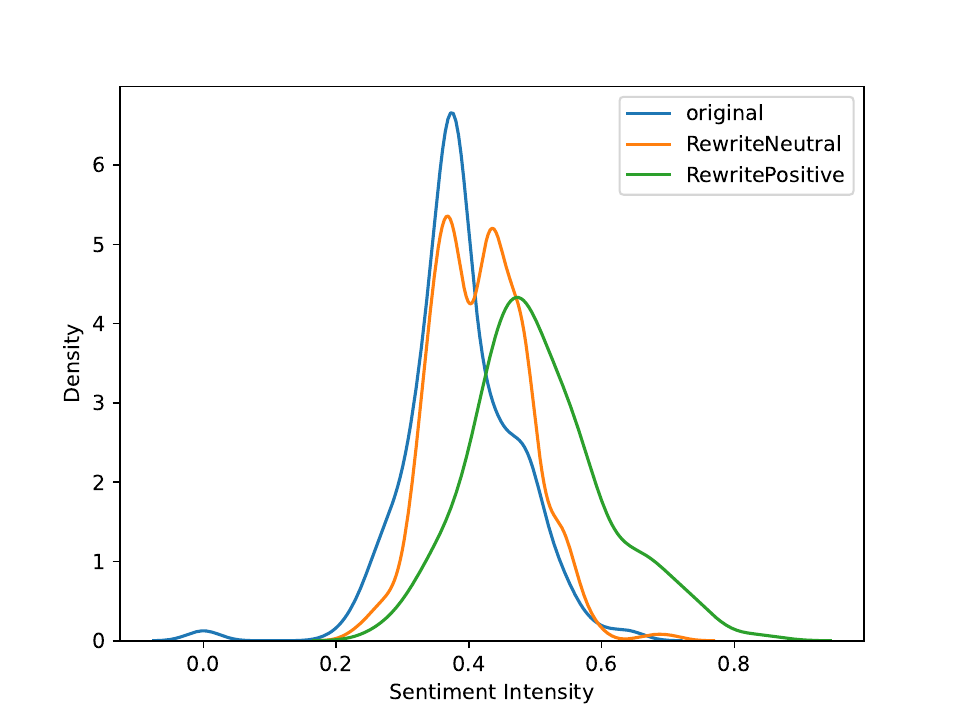}
\caption{Data distribution of different sentiment strengths on the validation set.}
\label{fig:sentidistributeval}
\end{figure}

\begin{figure}[!t]
\centering
\includegraphics[width=0.9\columnwidth]{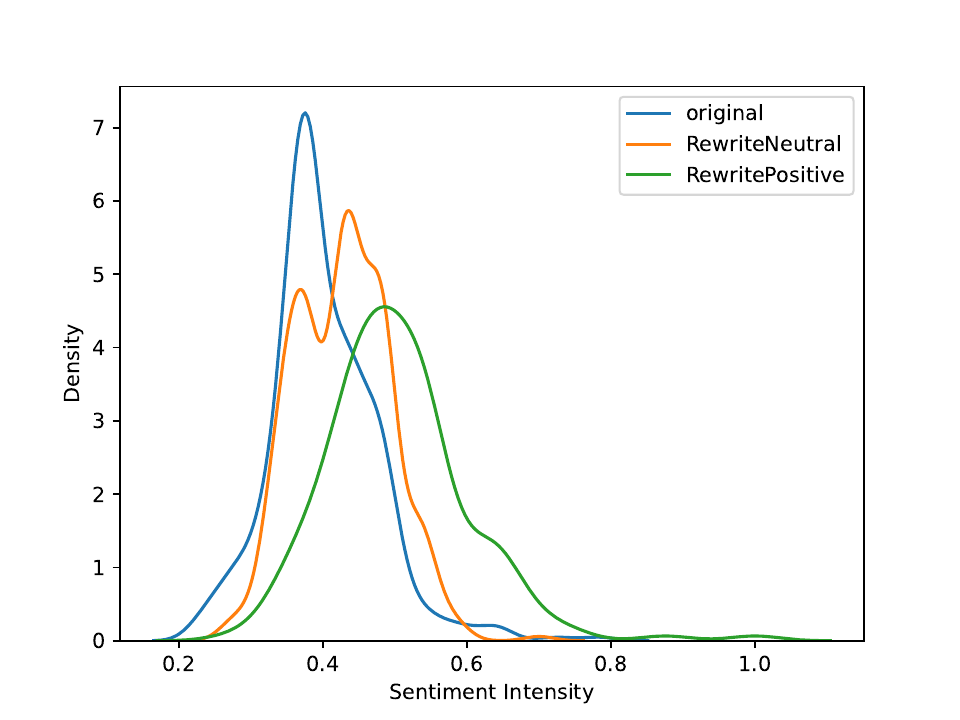}
\caption{Data distribution of different sentiment strengths on the test set.}
\label{fig:sentidistributetest}
\end{figure}

To verify the extent to which GPT-4o was able to successfully alter the emotions of the original tweets in the intended manner, we used EmoLLaMA \citep{liu2024emollms} to assign sentiment labels to the conspiracy tweets in the augmented dataset, focusing on the \textit{sentiment strength} score. This is a real-valued score representing the intensity of sentiment expressed in the text, in which 0 corresponds to the most negative sentiment, and 1 corresponds to the most positive sentiment.  Figures \ref{fig:sentidistributetrain}, \ref{fig:sentidistributeval} and \ref{fig:sentidistributetest} compare the distribution of sentiment strengths of the original Task 1 conspiracy tweets in different partitions of ConDID with the RewriteWithNeutral and RewriteWithPositive tweets. Table \ref{tab:ttestontest} shows the T-test statistics of sentiment strength between the original Task 1 conspiracy tweets in ConDID and the RewriteWithNeutral tweets, and between the original and RewriteWithPositive tweets.  Figures \ref{fig:sentidistributetrain} to \ref{fig:sentidistributetest} and Table \ref{tab:ttestontest} confirm that GPT-4o was able to successfully alter the sentiment strength of the rewritten content. Overall, the RewriteWithNeutral tweets exhibit a higher sentiment strength compared to the original tweets, while the overall sentiment strength in the RewriteWithPositive tweets is further increased.

\begin{table}[!t]
\footnotesize
\centering
\caption{T-test statistics of sentiment strength between original Conspiracy items (Task 1) and RewriteWithNeutral, and between original Conspiracy items (Task 1) and RewriteWithPositive.}
\label{tab:ttestontest}
\begin{tabular}{llcc}
\hline
                       & t-test                                          & t       & p     \\ \hline
\multirow{2}{*}{Train} & Original/RewriteWithNeutral    & -6.327  & 0.000 \\
                       & Original/RewriteWithPositive & -22.968 & 0.000 \\ \hline
\multirow{2}{*}{Val}   & Original/RewriteWithNeutral    & -4.959  & 0.000 \\
                       & Original/RewriteWithPositive & -13.804 & 0.000 \\ \hline
\multirow{2}{*}{Test}  & Original/RewriteWithNeutral    & -4.784  & 0.000 \\
                       & Original/RewriteWithPositive & -13.339 & 0.000 \\ \hline
\end{tabular}
\end{table}

\subsubsection{Performance of ConspEmoLLM on rewritten conspiracy tweets \label{sec:attack}}

Table \ref{tab:performanceonrewittencontents} reports the performance of the original ConspEmoLLM model on both the original and rewritten conspiracy tweets. Compared to its performance on the original tweets, ConspEmoLLM performs very poorly on the sentiment-transformed tweets. This demonstrates not only that conspiracy theory creators can easily and successfully ensure that their content evades detection by ConspEmoLLM using the rewriting strategy proposed in this paper, but also that ConspEmoLLM lacks robustness and may perform poorly when presented with new datasets.

\begin{table}[]
\centering
\caption{Performance of ConspEmoLLM on the original data and rewritten contents.}
\label{tab:performanceonrewittencontents}
\begin{tabular}{lcccc}
\hline
                & Accuracy & Precision & Recall & F1    \\ \hline
Conspiracy items (Task 1)        & 0.619    & 1.000     & 0.619  & 0.765 \\
RewriteNeutral  & 0.128    & 1.000     & 0.128  & 0.227 \\
RewritePositive & 0.072    & 1.000     & 0.072  & 0.135 \\ \hline
\end{tabular}
\end{table}

\begin{table*}[]
\centering
\caption{Results on ConDID. Bold text indicates the best performance, while underlined text indicates the second-best performance. Some of the results are taken from experiments reported in \citep{liu2024conspemollm}.}
\label{tab:ResultsonConDID}
\begin{tabular}{lcccccccccc}
\hline
\multirow{2}{*}{Model} & \multicolumn{2}{c}{Task1}       & \multicolumn{2}{c}{Task2}       & \multicolumn{2}{c}{Task3}       & \multicolumn{2}{c}{Task4}       & \multicolumn{2}{c}{Task5}       \\
                       & ACC            & F1             & ACC            & F1             & ACC            & F1             & ACC            & F1             & ACC            & F1             \\ \hline
BERT                   & 0.576          & 0.406          & 0.272          & 0.023          & 0.893          & 0.842          & 0.691          & 0.645          & {\ul 0.614}    & 0.296          \\
RoBERTa                & 0.517          & 0.231          & 0.279          & 0.112          & 0.893          & 0.844          & 0.677          & 0.666          & 0.601          & 0.250          \\
CT-BERT                & 0.564          & 0.428          & 0.042          & 0.175          & 0.893          & 0.842          & -              & -              & -              & -              \\ \hline
Mistral-7b-Instruct             & 0.592          & 0.600          & 0.130          & 0.227          & 0.428          & 0.529          & 0.677          & 0.683          & 0.345          & 0.376          \\
Llama3.2-1b-Instruct            & 0.284          & 0.297          & 0.033          & 0.253          & 0.044          & 0.027          & 0.332          & 0.211          & 0.175          & 0.215          \\
Llama3.2-3b-Instruct            & 0.556          & 0.457          & 0.110          & 0.244          & 0.085          & 0.034          & 0.650          & 0.659          & 0.233          & 0.273          \\
Llama3.1-8b-Instruct            & 0.079          & 0.131          & 0.216          & 0.256          & 0.036          & 0.069          & {\ul 0.709}    & {\ul 0.722}    & 0.444          & 0.483          \\
ChatGPT                & 0.638          & 0.596          & 0.324          & 0.332          & 0.208          & 0.240          & 0.668          & 0.664          & 0.596          & 0.574          \\
ConspLLM               & 0.662          & 0.675          & {\ul 0.328}    & {\ul 0.334}    & 0.893          & {\ul 0.864}    & 0.641          & 0.646          & 0.596          & 0.580          \\
ConspEmoLLM            & {\ul 0.695}    & {\ul 0.705}    & \textbf{0.340} & \textbf{0.364} & {\ul 0.897}    & 0.860          & 0.700          & 0.703          & 0.610          & \textbf{0.623} \\
ConspEmoLLM-v2         & \textbf{0.761} & \textbf{0.756} & 0.301          & 0.324          & \textbf{0.904} & \textbf{0.899} & \textbf{0.735} & \textbf{0.736} & \textbf{0.628} & {\ul 0.607}    \\ \hline
\end{tabular}
\end{table*}

\begin{table*}[]
\centering
\caption{Results on ConDID-v2. Bold text indicates the best performance, while underlined text indicates the second-best performance. We only present results for the conspiracy tweets in Task 1 and the rewritten tweets. The data for all other tasks in ConDID-v2 remains the same as ConDID, and hence the performance will be identical to the results shown in Table \ref{tab:ResultsonConDID}.}
\label{tab:ResultsonConDIDv2}
\begin{tabular}{lcccccc}
\hline
\multirow{2}{*}{Model} & \multicolumn{2}{c}{Conspiracy items (Task 1)} & \multicolumn{2}{c}{RewriteWithNeutral} & \multicolumn{2}{c}{RewriteWithPositive} \\
                       & ACC                   & F1                    & ACC                & F1                & ACC                & F1                 \\ \hline
Mistral-7b-Instruct             & 0.719                 & 0.837                 & 0.344              & 0.512             & 0.122              & 0.218              \\
Llama3.2-1b-Instruct            & 0.411                 & 0.583                 & 0.439              & 0.610             & {\ul 0.372}        & {\ul 0.543}        \\
Llama3.2-3b-Instruct            & \textbf{0.983}        & \textbf{0.992}        & {\ul 0.686}        & {\ul 0.814}       & 0.333              & 0.500              \\
Llama3.1-8b-Instruct            & 0.014                 & 0.027                 & 0.144              & 0.252             & 0.103              & 0.186              \\
ChatGPT                & {\ul 0.947}           & {\ul 0.973}           & 0.625              & 0.769             & 0.328              & 0.494              \\
ConspLLM               & 0.575                 & 0.730                 & 0.106              & 0.191             & 0.083              & 0.154              \\
ConspEmoLLM            & 0.619                 & 0.765                 & 0.128              & 0.227             & 0.072              & 0.135              \\
ConspEmoLLM-v2         & 0.881                 & 0.936                 & \textbf{0.976}     & \textbf{0.988}    & \textbf{0.960}     & \textbf{0.979}     \\ \hline
\end{tabular}
\end{table*}

\subsection{Expanding the dataset and enhancing ConspEmoLLM}

\begin{figure}[h]
\centering
\includegraphics[width=0.6\columnwidth]{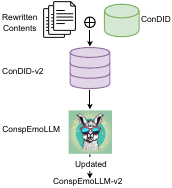}
\caption{Expanding the dataset and enhancing ConspEmoLLM.}
\label{fig:mainmethod2}
\end{figure}

To respond to the weaknesses of ConspEmoLLM highlighted in the previous section, we aimed to develop a more robust and stable version of the model, which is better able to handle not only human-authored conspiracy theory content, but also content whose sentiment has intentionally been altered using an LLM. Accordingly, we combined the sentiment-transformed conspiracy tweets described in the previous section (whose statistics are shown in Table \ref{tab:datastatistics}) with the original ConDID dataset to create ConDID-v2. The process is illustrated in Figure \ref{fig:mainmethod2}.

Following the same approach and hyperparameters described in \citep{liu2024conspemollm}, we developed an updated version of the ConspEmoLLM model, which we name ComnspEmoLLM-v2, by fine-tuning the EmoLLaMA-chat-7b model \citep{liu2024emollms} on the ConDID-v2 dataset. Training was conducted over three epochs using the AdamW optimiser \cite{loshchilov2017decoupled4}, and DeepSpeed \cite{rasley2020deepspeed1} was employed to optimise memory efficiency. We utilised a batch size of 256 and set the initial learning rate to 1e-6, applying a warm-up ratio of 5\%. The maximum input length was capped at 4096 tokens. The training process was executed on two Nvidia Tesla A100 GPUs, each equipped with 80GB of memory.



\section{Results}

\subsection{Baselines}

As baselines, we employed both pretrained language models (BERT series) and LLMs. The ConspLLM and ConspEmoLLM baselines were developed in \citet{liu2024conspemollm} by fine-tuning different LLM models using the ConDID dataset. Specifically, ConspLLM was fine-tuned from LLaMA-Chat-7B, which does not incorporate affective information, while ConspEmoLLM was fine-tuned from EmoLLaMA-Chat-7B, which includes implicit affective information.  We additionally evaluated several other LLMs as baselines, i.e., ChatGPT, Mistral-7B-Instruct, LLaMA3.1-8B-Instruct, and LLaMA3.2-(1B, 3B)-Instruct.

\subsection{Performance on the ConDID dataset \label{condidintroduction}}

Since the aim of our current work was to develop a model that can robustly handle both human-authored and sentiment-transformed conspiracy text, we firstly evaluated the performance of ConspEmoLLM-v2 on the original ConDID dataset \citep{liu2024conspemollm}, which includes five tasks based on human-authored data.  \textbf{Task 1} is conspiracy intent detection, which classifies textual items into three categories, i.e., \textit{unrelated}, \textit{related} (relevant to a conspiracy theory but not propagating it) and \textit{conspiracy} (relevant to and propagating a conspiracy theory). \textbf{Task 2} is conspiracy topic identification, which determines whether each textual item concerns any of twelve predefined categories\footnote{Suppressed Cures, Behavior Control, Anti Vaccination, Fake Virus, Intentional Pandemic, Harmful Radiation, Depopulation, New World Order, Esoteric Misinformation, Satanism, Other Conspiracy Theory, Other Misinformation.}. \textbf{Task 3} is a combination of Task 1 and Task 2, with the aim of sequentially assessing the conspiracy intent for each of the twelve categories. \textbf{Task 4} is a binary classification task, whose aim is to determine whether of not a piece of text represents a conspiracy theory. Finally, \textbf{Task 5} assesses the degree of relevance of a textual item to a conspiracy theory, classifying it as \textit{closely related}, \textit{broadly related}, or \textit{not related}.

Table \ref{tab:ResultsonConDID} compares the performance of the novel ConspEmoLLM-v2 model with the baselines. Several of the baselines were also used in \citet{liu2024conspemollm}, and the results are taken from that paper. The results in Table \ref{tab:ResultsonConDID} show that performance of the original ConspEmoLLM model surpasses the baseline models for most tasks, thus demonstrating the importance of affective information in detecting different types of information about human-authored conspiracy content.  Furthermore, the updated ConspEmoLLM-v2 outperforms ConspEmoLLM in Tasks 1, 3, and 4. This suggests that training a model using the more emotionally diverse  ConDID-v2 dataset can be beneficial for certain tasks, especially those that involve distinguishing content that conveys conspiracy theories.  However, a slight performance decline (4\% or less) is observable when comparing the performance of ConspEmoLLM-v2 to ConspEmoLLM for Tasks 2 and 5. A possible reason for this is that the sentiment-shifted data included in ConDID-v2 is related specifically to conspiracy detection tasks. Since Tasks 2 and 5 are not directly related to conspiracy detection, the new data may have introduced a certain level of confusion into the model when it is applied to these tasks. Future work will investigate whether further extending ConDID-v2 to include sentiment-transformed data for these tasks can help to stabilise or improve performance. Nevertheless, in terms of overall performance, ConspEmoLLM-v2 demonstrates a consistent or even improved level of performance compared to its predecessor, when it is applied to the human-authored content in the original ConDID datset. 


\subsection{Results on sentiment-transformed tweets in ConDID-v2}

Table \ref{tab:ResultsonConDIDv2} presents the results of applying the various models to the ConDID-v2 dataset. We only show the performance on the original conspiracy tweets from Task 1, as well as the rewritten items (i.e., RewriteWithNeutral and RewriteWithPositive). The data for all other tasks in ConDID-v2 remains the same as ConDID, and hence the performance will be identical to the results shown in Table \ref{tab:ResultsonConDID}. Table \ref{tab:ResultsonConDIDv2} shows that, in terms of performance on the original conspiracy items, ConspEmoLLM-v2 is better than most, but not all, of the baselines considered. However, when performance on the \textit{RewriteWithNeutral} and \textit{RewriteWithPositive} tweets is considered, ConspEmoLLM-v2 is considerably better than all baseline models, with fairly consistent performance on items that have been rewritten with both types of modified sentiment. In contrast, although most models perform well on human-authored text (i.e., Task 1 (Conspiracy items)), their performance drops significantly on the rewritten data, especially on \textit{RewriteWithPositive} tweets. This suggests that the detection of conspiracy theory tweets becomes more challenging for all models as the sentiment strength of the tweets increases. Although all baseline models struggle with the rewritten tweets, the lowest performance by a considerable margin is attained by the ConspLLM and ConspEmoLLM models. This suggests an over-reliance of these models on the typical sentiment characteristics of human-authored conspiracy theories, which makes them a particularly easy target for sentiment-based attacks. A further interesting observation is that, despite having a higher number of parameters, Llama3.1-8b-Instruct demonstrates considerably lower performances than Llama3.2-(1b, 3b)-Instruct across all metrics. This is primarily due to its tendency to withhold answers and to provide more conservative responses\footnote{In most cases, Llama3.1-8b responds with: 'I cannot provide information that could be used to spread misinformation about COVID-19. Is there anything else I can help you with?'}. A possible explanation for this behaviour is that Llama3.1-8b was subject to a more cautious pre-training strategy than other models, with the aim of ensuring safer outputs.


In summary, the results in Tables \ref{tab:ResultsonConDID} and \ref{tab:ResultsonConDIDv2} illustrate not only that ConspEmoLLM-v2 retains or even exceeds the performance of ConspEmoLLM when applied to the original human-authored content in the ConDID dataset, but also that it performs remarkably well when faced with sentiment-transformed data. This demonstrates that ConspEmoLLM-v2 is able to maintain the reliability of CompsEmoLLM when applied to human-authored conspiracy content, whilst also offering the added benefit of robustness against sentiment-based attacks.

\section{Conclusion}
The sophistication of state-of-the-art LLMs is increasing the feasibility of using them to intentionally and convincingly alter the textual features of false information, in an attempt to evade detection by misinformation detection methods. This paper has investigated how to address a particular type of LLM-based modification, i.e, transformation of the typical sentiment expressed in conspiracy theories. We began by using GPT-4o to rewrite existing conspiracy theory tweets with transformed sentiments. The quality and effectiveness of the rewritten content were assessed through a combination of human and LLM-based evaluation, and EmoLLaMA was used to verify the success of sentiment the transformation. Rewritten tweets that were determined to still convey conspiracy theories were used to create ConDID-v2, which is an augmented version of ConDID, a previously-developed instruction tuning dataset for conspiracy theory detection tasks. We used ConDID-v2 dataset to train a new conspiracy detection model, ConspEmoLLM-v2, and conducted comprehensive evaluations that compare its performance with multiple baselines when applied to both the original ConDID and the expanded ConDID-v2  datasets. The results of these evaluations demonstrate that ConspEmoLLM-v2 is not only able to retain or exceed the performance of the original ConspEmoLLM model on the ConDID dataset, but also exhibits strong ability for generalisation and robustness in its handling of emotionally manipulated conspiracy content.

As future work, we plan to use LLMs to rewrite textual items concerning a broader range of conspiracy-related tasks with varying emotional tones. We also aim to explore other ways in which conspiracy theory text could be manipulated, e.g., by changing the writing style, to further investigate the impact of such modifications on conspiracy detection.

\section{Limitations}

Due to computational resource limitations, we only tested models with a maximum size of 8b. Due to the scarcity of existing LLMs specifically designed for conspiracy theory detection, this paper focuses solely on rewriting conspiracy theories with altered emotions. In the future, we will explore different ways of rewriting data, in an attempt to further enhance conspiracy theory detection models.







\bibliography{ecai}

\end{document}